\documentclass[runningheads]{llncs}
\usepackage{amsmath, amssymb}
\usepackage[T1]{fontenc}
\usepackage{graphicx}
\usepackage{hyperref}
\usepackage{enumitem} 
\usepackage{subfigure}
\begin{document}
\title{Optimal Take-off under Fuzzy Clearances}
%
%
\author{Hugo Henry\inst{1}\orcidID{0009-0001-2530-3521} \and Arthur Tsai\inst{2}\orcidID{0009-0009-0002-4735} \and Dr. Kelly Cohen\inst{3}\orcidID{0000-0002-8655-1465}}
\authorrunning{H. Henry et al.}
%
\institute{University of Cincinnati, Cincinnati OH 45221, USA\\
 \email{henryho@mail.uc.edu}\\ \email{tsaiti@mail.uc.edu}\\
\email{cohenky@ucmail.uc.edu}}
\maketitle              
%


\begin{abstract}
This paper presents a hybrid obstacle-avoidance architecture that integrates Optimal Control under clearance with a Fuzzy Rule-Based System (FRBS) to enable adaptive constraint handling for unmanned aircraft. Motivated by the limitations of classical optimal control under uncertainty and the need for interpretable decision-making in safety-critical aviation systems, we design a three-stage Takagi–Sugeno– Kang fuzzy layer that modulates constraint radii, urgency levels, and activation decisions based on regulatory separation minima and airworthiness guidelines from FAA and EASA. These fuzzy-derived clearances are then incorporated as soft constraints into an optimal control problem solved using the FALCON toolbox and IPOPT. The framework aims to reduce unnecessary recomputations by selectively activating obstacle-avoidance updates while maintaining compliance with aviation procedures.
A proof-of-concept implementation using a simplified aircraft model demonstrates that the approach can generate optimal trajectories with computation times of 2–3 seconds per iteration in a single-threaded MATLAB environment, suggesting feasibility for near real-time applications. However, our experiments revealed a critical software incompatibility in the latest versions of FALCON and IPOPT, in which the Lagrangian penalty term remained identically zero, preventing proper constraint enforcement. This behavior was consistent across scenarios and indicates a solver–toolbox regression rather than a modeling flaw. Future work includes validating this effect by reverting to earlier software versions, optimizing the fuzzy membership functions using evolutionary methods, and extending the system to higher-fidelity aircraft models and stochastic obstacle environments.

\end{abstract}

\section{Introduction}
Control systems have been in the work for decades, especially in complex engineering domains such as flight mechanics, with fly-by-wire implemented since the 1970s for the military and 1988 for civil aircraft. Those control systems encompass every domain, such as aerodynamics with surface control or thermodynamics and fluid flow, with the example of the FADEC (Full Authority Digital Engine Control). As the distribution of automated systems grows in order to reduce the pilot's load, the question of efficiency arises, for that multiple applications have been developed, such as linear regulators, and then $H^\infty$, while finally arriving at the notion of Optimal Control \cite{hull}. However, this optimality is very fragile to uncertainty and noise; therefore, the notion of optimality under clearance has been created to maintain safety and bounds to the system \cite{ben asher}. But clearances and constraints may not be absolute or have varying degrees of importance to the control system, depending on time. Therefore, we wished to explore the combination of fuzzy decision making for adaptive constraints and creating an equivalent of a detect and avoid system while assuming perfect detection conditions (that assumption is detailed in part \ref{section:constraint def}). In this paper, we will explore what happens when we couple the world of optimal control with fuzzy-based clearances and constraints. The choice of a Fuzzy system as the decision layer was made due to the airworthiness for autonomous airborne systems \cite{hugo}. The use case presented in this research is a take-off situation with varying number of obstacles, the methodology presented is a global control technique but for the sake of memory usage, we opted to look at a short time problem.

\section{Methodology}

For this paper, we base the theory on the book published by Joseph Z. Ben Asher et al. \cite{ben asher} and apply a dynamic fuzzy system in tandem with the Optimal control offline solver Falcon.m made by TUM \cite{Falcon}.
The creation of the Fuzzy Rule-Based System (FRBS) is based on actual Airworthiness Directives (AD/SB), Advisory circulars (AC), and other regulations from the authorities, mainly the Federal Aviation Agency (FAA) and the European Union Aviation Safety Agency (EASA). The method of validation of the model will be purely theoretical, given the low-fidelity aircraft model used to design the system (model given by the Falcon.m library, lightly modified for constraints) \cite{Falcon}. 

\section{Optimal Control under Clearance}
\label{section:1}
Consider an UAV taking off and climbing to its desired cruise altitude by following an optimal path. This optimal path represents the safest, smoothest, and most cost-efficient route to reach a target destination under ideal conditions. However, the sky is not always empty. Other air vehicles may share the same airspace, and flocks of birds may appear unexpectedly. Assuming that these obstacles are unaware of the presence of our UAV, and that our UAV is equipped with perfect radar, it is our responsibility to ensure deconfliction with these obstacles, maintain safe separation from them, and guarantee absolute safety. This nonlinear clearance problem in continuous time can be transcribed to a finite-dimensional optimization problem.
There are multiple approaches that could be implemented to solve this optimal control clearance problem. One option would be to employ a receding horizon optimal control strategy to generate a collision free trajectory. Another is to predict the future position of the obstacle using a Kalman filter based static solver to better account for uncertain obstacle motion. The third option is the combination of these two methods. However, the combined approach is expected to be significantly more computationally intensive, which may limit its suitability for real-time onboard implementation. 
In order to solve this problem, we use the Falcon.m \cite{Falcon} toolbox from TUM, a static solver that compute one configuration (phase) at a time. This allow us to solve the problem pointwise by operating on a sequence of phases with static constraints. However, it is still not a time reactive dynamic solver, which can create conflicts with obstacles juxtaposed near the endpoints of phases. In contrast, our objective is a time dependent obstacle avoidance system that regularly updates obstacle positions and danger zones. Therefore, the states of moving obstacles must be updated at each time step, and the optimal trajectory must be regenerated accordingly to ensure continuous safety, proper deconfliction, and adherence to the required clearance margins throughout the flight. This poses a new challenge. The system may repeatedly rerun the optimization and regenerate the same trajectory even when no new obstacles are present or when the previously added constraint is no longer relevant, resulting in unnecessary computational effort and redundant recalculations.
To further address this issue and improve the overall efficiency and robustness of the system, a fuzzy-logic-based decision-making layer is introduced in the next section.

\section{Fuzzy Clearances}
To make the system more efficient and robust, we introduce a first-order Takagi-Sugeno-Kang (TSK) FRBS. The fuzzy inference system evaluates the parameters of each detected obstacle given by the perfect radar and determines whether it should be treated as an active constraint in the optimal control problem through a three-stage process: fuzzification, rule-based inference, and defuzzification. The rule base is designed in accordance with existing aviation regulations to ensure that all decisions are operationally meaningful and compliant with safety standards. By embedding expert knowledge into its rule base, the fuzzy layer provides graded and interpretable \cite{Lynn} decisions regarding threat severity and required clearance levels. This design ensures that the overall system remains flexible, resilient to uncertainty, and capable of intelligently adapting its constraints before the optimal solver recomputes a safe trajectory. It should be noted that the membership functions presented in the following section have not been optimized and are therefore intended to serve as a hot start for subsequent optimization processes, such as Genetic Algorithms (GA).

\subsection{Application of Fuzzy Sets to constraints}
\label{section:constraint def}

\begin{figure}
    \centering
    \begin{minipage}{.49\textwidth}
    \centering
    \includegraphics[width =1\textwidth]{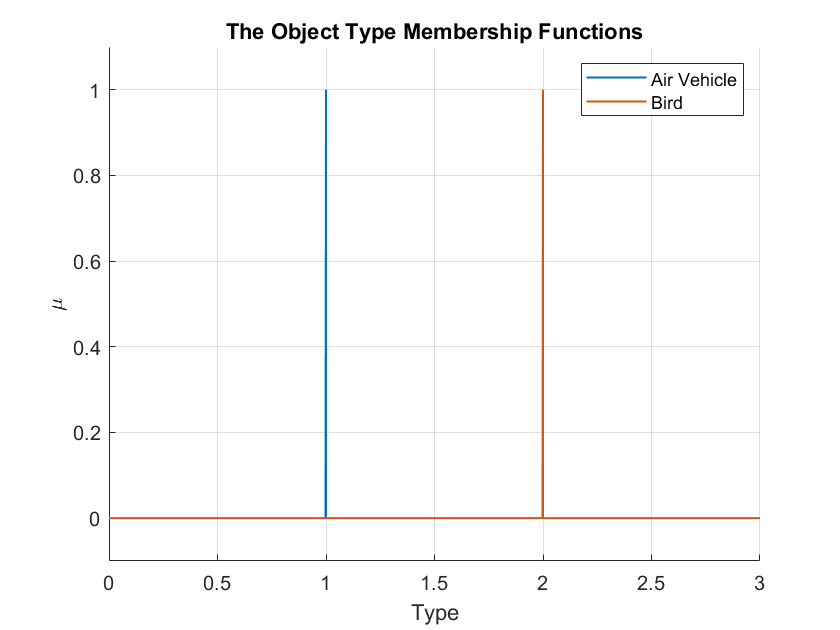}
    \caption{Object Type Membership}
    \label{fig:objecttype}
    \end{minipage}
    \begin{minipage}{.49\textwidth}
    \centering
    \includegraphics[width = 1\textwidth]{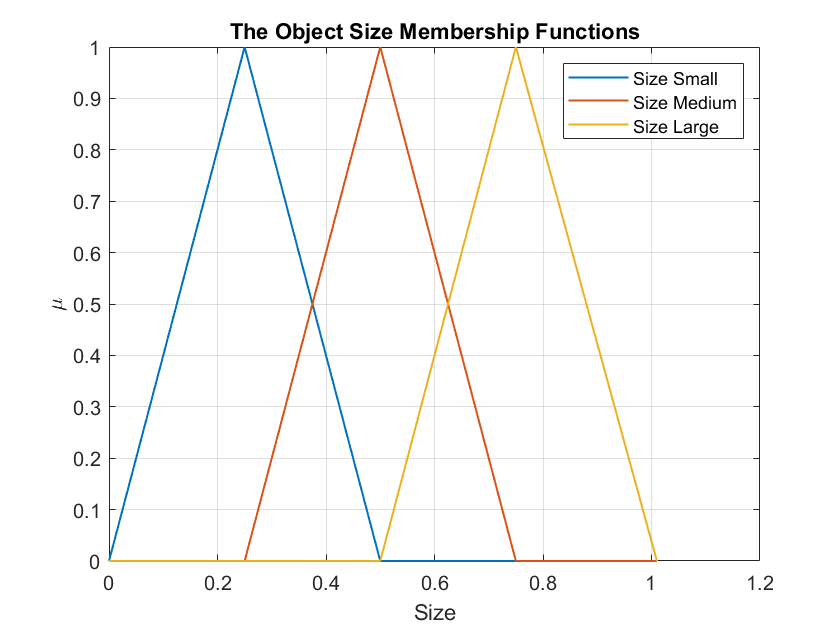}
    \caption{Object Size Membership}
    \label{fig:objectsize}
    \end{minipage}
    
\end{figure}

For this specific problem setup, four parameters of each detected object are normalized and used as inputs to the fuzzy inference system: type, size $S_{i}$, position in the global frame, and velocity in the global frame. The fuzzy inference architecture consists of three fuzzy subsystems, each with two inputs and one output that determines, respectively, the radius of the constraint $R_{i}$, the level of urgency $U_{i}$, and the activation of optimal path recalculation.
The inputs to the \textit{radius of the constraint} fuzzy subsystem are the object type and the object size. The \textit{type} input has two membership functions: \textit{air vehicle} and \textit{bird} (Figure \ref{fig:objecttype}). The \textit{size} input has three membership functions: \textit{Small}, \textit{Medium}, and \textit{Large}  (Figure \ref{fig:objectsize}). The rule-based reasoning of this subsystem is inspired by air traffic management guidelines and operational judgment, and can be described as follows, for an Air vehicle: vertical separation, obtained by assigning different levels selected from the table of cruising levels in Appendix 3 to the Annex to Implementing Regulation (EU) No 923/2012 \cite{EEA}, except that the correlation of levels to track as prescribed therein shall not apply whenever otherwise indicated in appropriate aeronautical information publications or ATC clearances. The constraints will appear not closer than 1000 m from the end of the runway, given that the aircraft is subject to the separation minima under ATS surveillance systems during take off. In a real airport setting, the transition time is usually around 15 to 30 seconds which at take off speed are equivalent to 2-3 kms, therefore by taking 1km, we ensure that we are in the specifications. The vertical separation minimum shall be a nominal 300 m (1,000 ft) up to and including FL-410 and a nominal 600 m (2,000 ft) above that level. Geometric height information shall not be used to establish vertical separation, moreover, the horizontal separation needs to be at least $ 3 nautical mile = 5556 m$ \cite{ATM}
Birds: given by the FAA \cite{avian}, we know that a specialized avian radar is capable of detecting a single duck-sized target at a maximum radius of 6 km and can follow a total of 1000 targets and segregate them if they are more than $50m$ from each other. 
Therefore, let's consider the following bird flock size range :
\begin{itemize}
    \item Let us consider that each bird is a perfect closed ball in $\mathbb{R}^3$ of radius $R>\epsilon>1$ so that $\bigcup_{i=1}^nB_i\subset B_R$ pairwise disjoint.
    \item  Let us assume a Kepler maximum density repartition in $\mathbb{R}^3$ with $\phi_{max}=\frac{3}{2}\pi\approx0.74$ so that $R\geq25(\frac{1000}{\phi_{max}})^{1/3}=277m$
\end{itemize}
Under these approximations, we can say that the bird detection size range would be a ball of radius $277\geq\epsilon\geq 1$ at a maximum of 6 km ahead.
The four rules of the radius of the constraint fuzzy set are summarized as follows :

\begin{enumerate}[label=(\arabic*), noitemsep]
    \item \textit{If Type} is \textit{air vehicle} and $S_{i}$ is \textit{Small}, \textit{Medium}, or \textit{Large}, then the radius of $R_{i}=$ 5556 m
    \item \textit{If Type} is \textit{bird} and $S_{i}$ is \textit{Small}, then the radius of $R_{i} = 2.5*S_{i} + 100$ $m$
    \item \textit{If Type} is \textit{bird} and $S_{i}$ is \textit{Medium}, then the radius of $R_{i} = 2.5*S_{i} + 200$ $m$
    \item \textit{If Type} is \textit{bird} and $S_{i}$ is \textit{Large}, then the radius of $R_{i} = 2.5*S_{i} + 300$ $m$
\end{enumerate}

\begin{figure}
    \begin{minipage}{.49\textwidth}
    \centering
    \includegraphics[width = 1\textwidth]{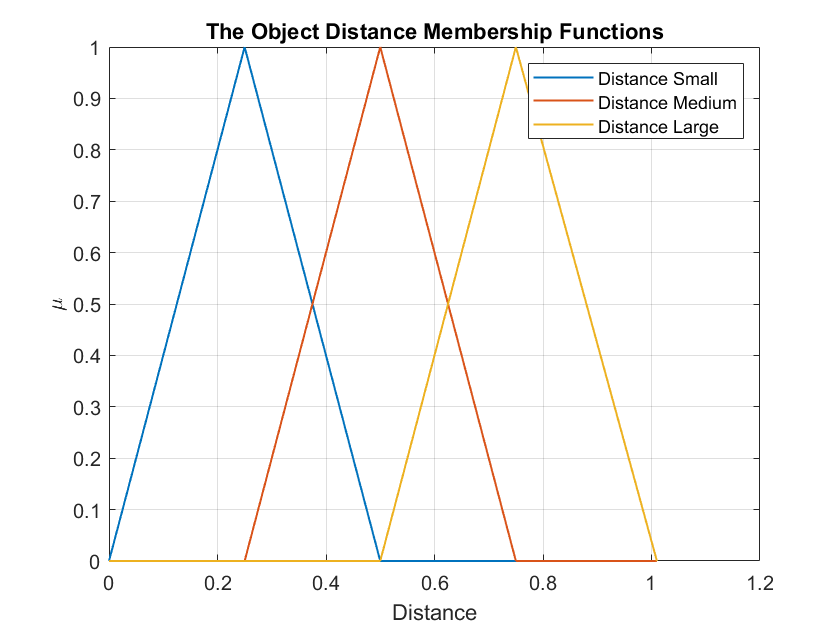}
    \caption{Object Distance Membership}
    \label{fig:objectdist}
    \end{minipage}
    \centering
    \begin{minipage}{.49\textwidth}
    \centering
    \includegraphics[width =1\textwidth]{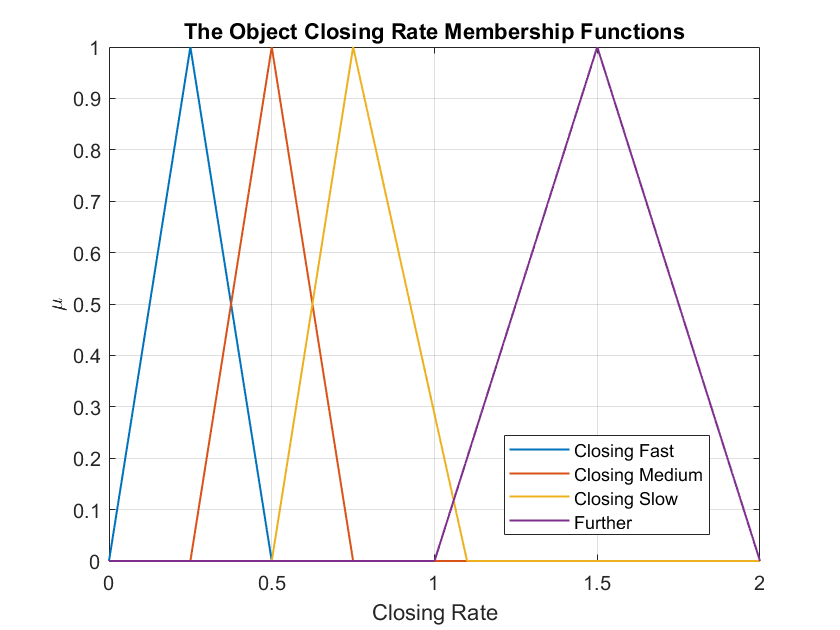}
    \caption{Object Closing-rate Membership}
    \label{fig:objectCR}
    \end{minipage}
    
    
\end{figure}

\begin{figure}
    \centering
    \begin{minipage}{.49\textwidth}
    \centering
    \includegraphics[width =0.75\textwidth]{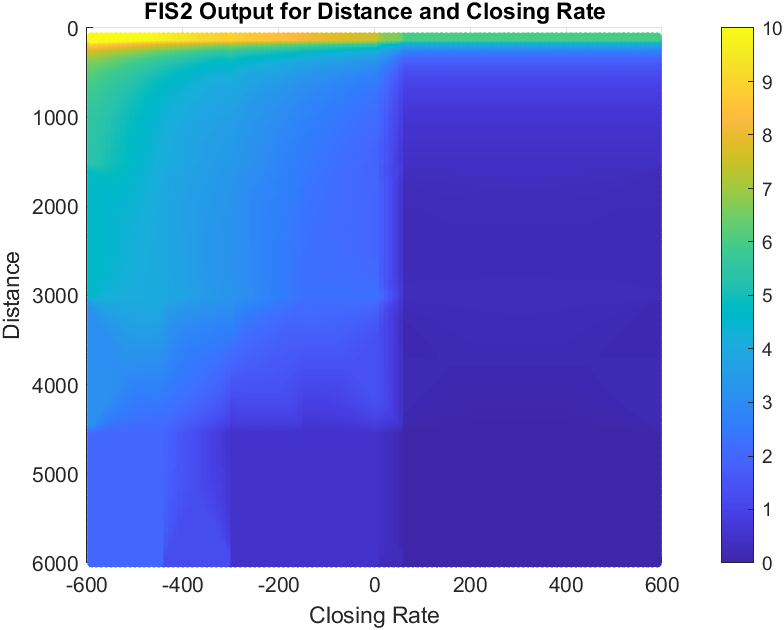}
    \caption{Urgency control surface}
    \label{fig:CSU}
    \end{minipage}
    \begin{minipage}{.49\textwidth}
    \centering
    \includegraphics[width =0.75\textwidth]{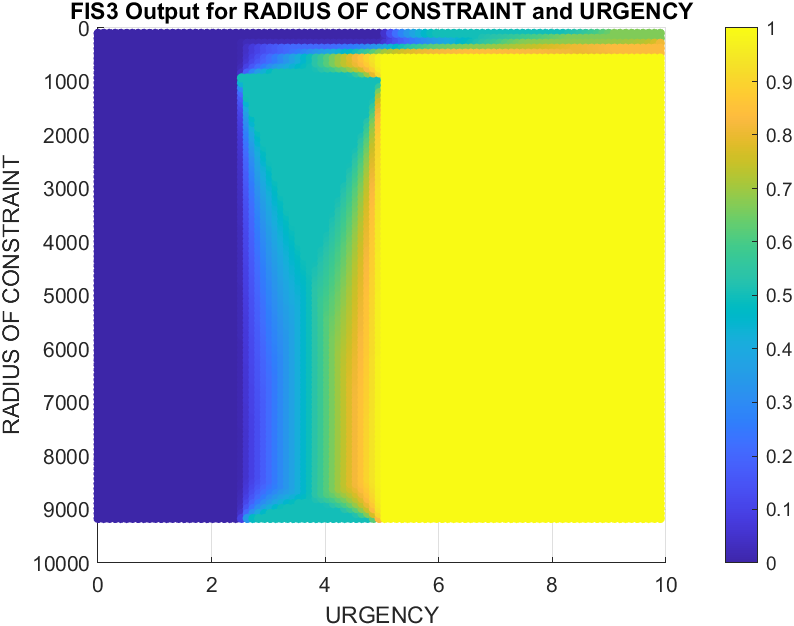}
    \caption{Activation control surface}
    \label{fig:CSA}
    \end{minipage}
\end{figure}

\noindent
The inputs to the urgency fuzzy subsystem are the distance and the closing rate between the UAV and a detected object. The distance $D$ is computed from the relative position vector $R_{p}$, while the closing rate $CR$ is obtained from the dot product of the relative position vector $R_{p}$ and the relative velocity vector $R_{v}$, normalized by the distance. The \textit{distance} variable is described using three membership functions: \textit{Small}, \textit{Medium}, and \textit{Large} (Figure \ref{fig:objectdist}). The \textit{closing rate} is represented by four membership functions: \textit{Further}, \textit{Closing Slow}, \textit{Closing Medium}, and \textit{Closing Fast} (Figure \ref{fig:objectCR}). The control surfaces of the Urgency Fig. \ref{fig:CSU} and constraint Activation Fig. \ref{fig:CSA}. From these control surfaces, it is evident that the Activation is non-monotonic and therefore requires refinement through an optimization process, using either gradient-based or evolutionary methods.

\begin{center}
    UAV position: $P_{0}=[x_{0},y_{0},z_{0}]$
\end{center} 
\begin{center}
    Detected object position: $P_{i}=[x_{i},y_{i},z_{i}]$
\end{center} 
\begin{equation}
    \centering R_{Pi}= [x_{i}-x_{0},y_{i}-y_{0},z_{i}-z_{0}]
\end{equation}
\begin{equation}
    \centering D_{i}= \sqrt{(x_{i}-x_{0})^2+(y_{i}-y_{0})^2+(z_{i}-z_{0})^2}
\end{equation}
\begin{center}
    UAV velocity: $V_{0}=[u_{0},v_{0},w_{0}]$
\end{center} 
\begin{equation}
    \centering R_{Vi}= [u_{i}-u_{0},v_{i}-v_{0},w_{i}-w_{0}]
\end{equation}
\begin{equation}
    \centering CR_{i}=\frac{R_{Pi}\cdot R_{Vi}}{D_{i}}
\end{equation}
\noindent
The four rules of the radius of the constraint fuzzy set are summarized as follows:

\begin{enumerate}[label=(\arabic*), noitemsep]
    \item \textit{If} $D_{i}$ is \textit{Large} and $CR_{i}$ is \textit{Further}, then $U_{i}=0$ 
    \item \textit{If} $D_{i}$ is \textit{Large} and $CR_{i}$ is \textit{Closing Slow}, then $U_{i}=0.5$  
    \item \textit{If} $D_{i}$ is \textit{Large} and $CR_{i}$ is \textit{Closing Medium}, then $U_{i}=0.5$  
    \item \textit{If} $D_{i}$ is \textit{Large} and $CR_{i}$ is \textit{Closing Fast}, then $U_{i}=2$  
    \item \textit{If} $D_{i}$ is \textit{Medium} and $CR_{i}$ is \textit{Further}, then $U_{i}=0.5*D_{i}$  
    \item \textit{If} $D_{i}$ is \textit{Medium} and $CR_{i}$ is \textit{Closing Slow}, then $U_{i}=0.5*D_{i}+2$   
    \item \textit{If} $D_{i}$ is \textit{Medium} and $CR_{i}$ is \textit{Closing Medium}, then $U_{i}=0.5*D_{i}+3$  
    \item \textit{If} $D_{i}$ is \textit{Medium} and $CR_{i}$ is \textit{Closing Fast}, then $U_{i}=0.5*D_{i}+4$    
    \item \textit{If} $D_{i}$ is \textit{Small} and $CR_{i}$ is \textit{Further}, then $U_{i}=\frac{0.1}{D_{i}}+1.5$ 
    \item \textit{If} $D_{i}$ is \textit{Small} and $CR_{i}$ is \textit{Closing Slow}, then $U_{i}=\frac{0.1}{D_{i}}-2.5*CR_{i}+4$ 
    \item \textit{If} $D_{i}$ is \textit{Small} and $CR_{i}$ is \textit{Closing Medium}, then $U_{i}=\frac{0.1}{D_{i}}-3*CR_{i}+4.5$
    \item \textit{If} $D_{i}$ is \textit{Small} and $CR_{i}$ is \textit{Closing Fast}, then $U_{i}=\frac{0.1}{D_{i}}-5*CR_{i}+5$  
\end{enumerate}

\begin{figure}
    \centering
    \begin{minipage}{.49\textwidth}
    \centering
    \includegraphics[width =0.75\textwidth]{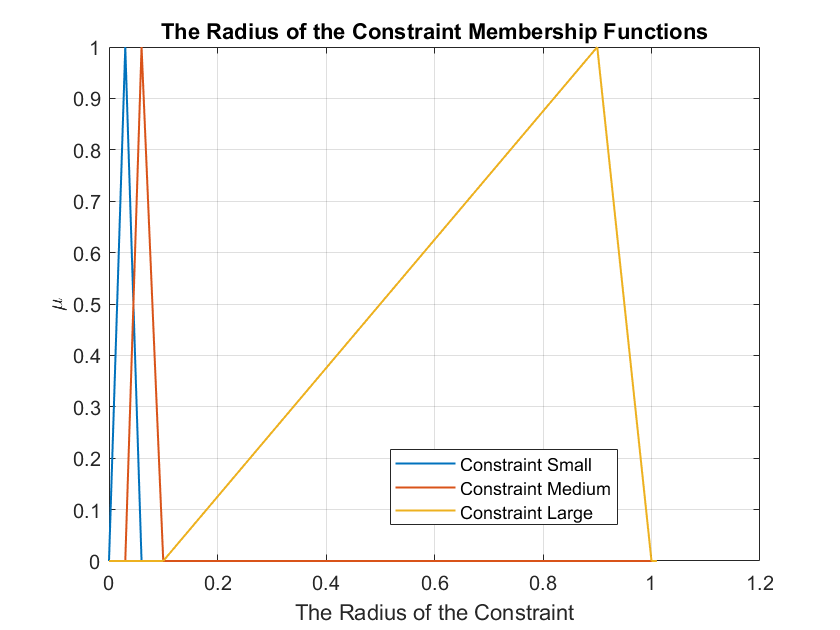}
    \caption{Constraint Radius Membership}
    \label{fig:constraintrad}
    \end{minipage}
    \begin{minipage}{.49\textwidth}
    \centering
    \includegraphics[width = 0.75\textwidth]{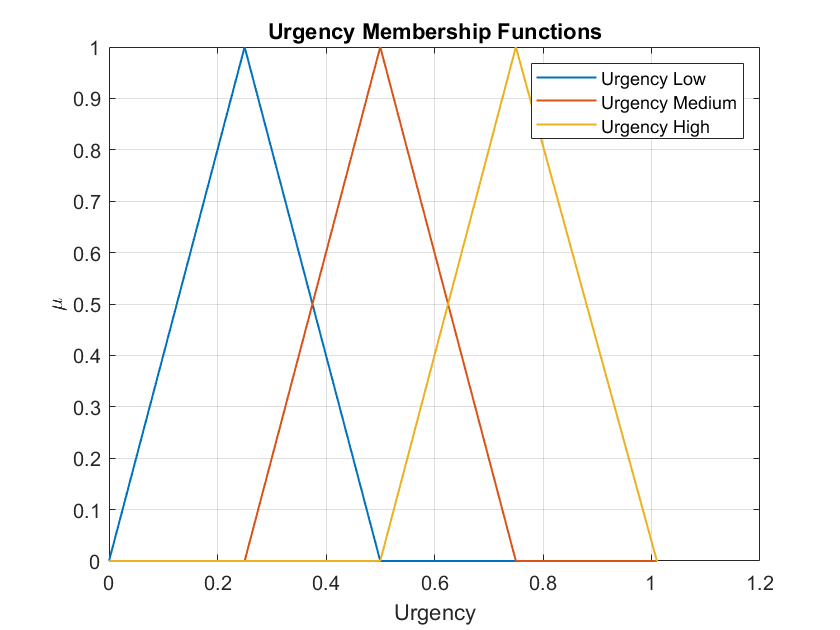}
    \caption{Urgency Membership}
    \label{fig:urgency}
    \end{minipage}
    
\end{figure}

\noindent
The inputs to the activation fuzzy subsystem are the outputs of the previous fuzzy subsystems, $R_{i}$ and $U_{i}$. The \textit{radius} input is described by three membership functions: \textit{Small}, \textit{Medium}, and \textit{Large} (Figure \ref{fig:constraintrad}). Similarly, the \textit{urgency} input is described by three membership functions: \textit{Low}, \textit{Medium}, and \textit{High} (Figure \ref{fig:urgency}). The rules of the activation fuzzy subsystem are summarized as follows:

\begin{enumerate}[label=(\arabic*), noitemsep]
    \item \textit{If} $R_{i}$ is \textit{Small}  and $U_{i}$ is \textit{Low}, then the \textit{activation} is 0
    \item \textit{If} $R_{i}$ is \textit{Small}  and $U_{i}$ is \textit{Medium}, then the \textit{activation} is 0
    \item \textit{If} $R_{i}$ is \textit{Small} and $U_{i}$ is \textit{High}, then the \textit{activation} is 1
    \item \textit{If} $R_{i}$ is \textit{Medium} and $U_{i}$ is \textit{Low}, then the \textit{activation} is 0
    \item \textit{If} $R_{i}$ is \textit{Medium} and $U_{i}$ is \textit{Medium}, then the \textit{activation} is 1
    \item \textit{If} $R_{i}$ is \textit{Medium} and $U_{i}$ is \textit{High}, then the \textit{activation} is 1
    \item \textit{If} $R_{i}$ is \textit{Large} and $U_{i}$ is \textit{Low}, then the \textit{activation} is 0
    \item \textit{If} $R_{i}$ is \textit{Large} and $U_{i}$ is \textit{Medium}, then the \textit{activation} is 1
    \item \textit{If} $R_{i}$ is \textit{Large}  and $U_{i}$ is \textit{High}, then the \textit{activation} is 1
\end{enumerate}

\subsection{Adaptation for optimal control}

\begin{figure}
    \centering
    \includegraphics[width=0.7\linewidth]{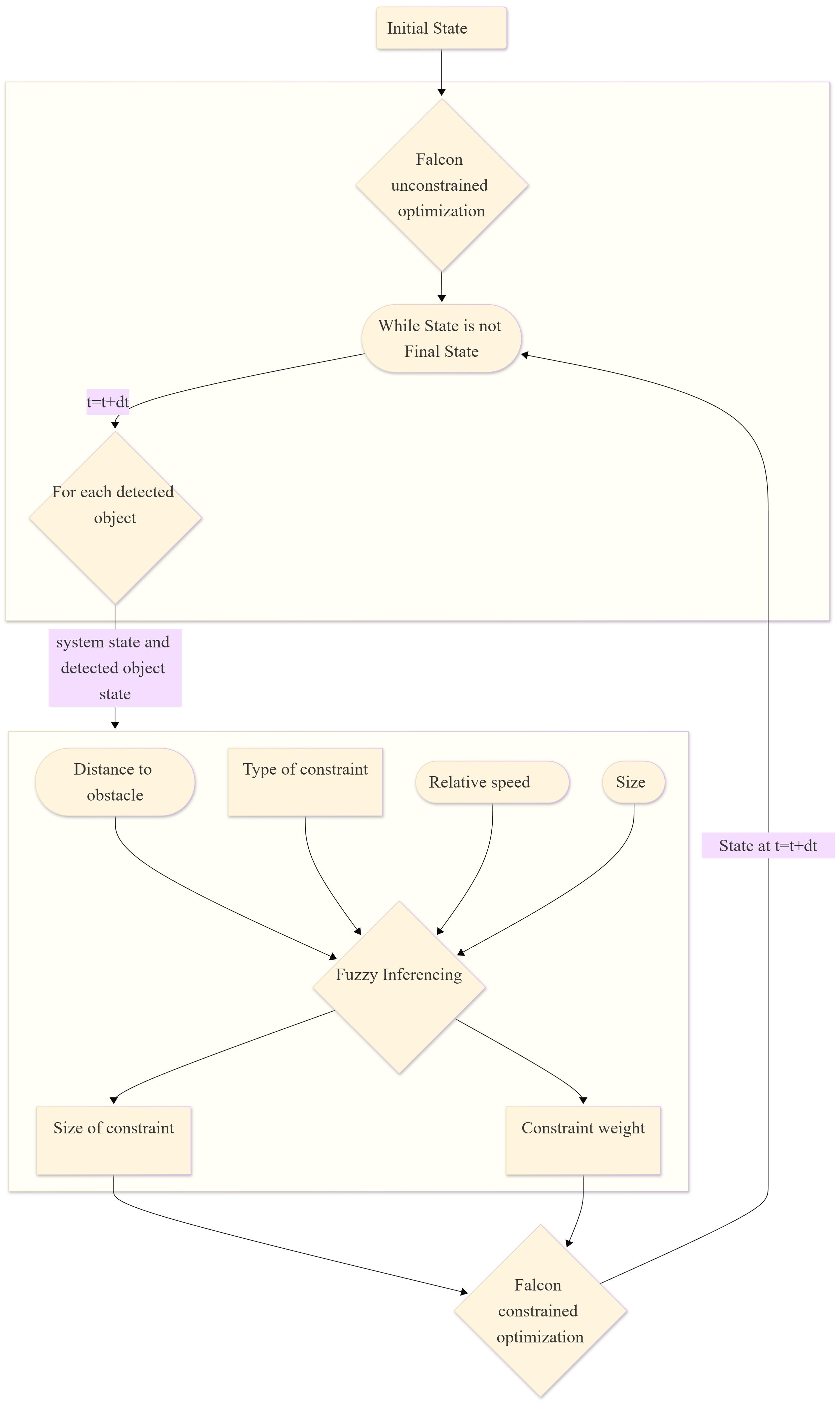}
    \caption{Path optimization Flowchart}
    \label{fig:flowchart}
\end{figure}

In the process of using FRBS to reduce the computational load of recalculation of the path and add dynamism to the constraints, we use the Falcon.m toolbox as said in part \ref{section:1}, and consider the following costs: 
\begin{itemize}
    \item A linear cost toward the final time
    \item A Lagrangian penalty encompassing all the violations of the constraints determined by the FIS
\end{itemize}
In order for the problem to be solvable, a Lagrangian cost as been chosen to represent the soft constraint, which, as opposed to a hard constraint, allows violations with a cost. Indeed, if hard constraints were implemented, the solver might generate a trajectory, but the update of the constraint might make the initial position of the problem formulation immediately juxtaposed to said constraint, making the problem unsolvable. Moreover, we consider no-go zones according to the Air traffic management guidelines, which are crisp changes in distancing before and after specific points. One of those being the Radar enabled zone addressed in part \ref{section:constraint def}. We did not try to implement those change in definitions which could be the object of another FIS. As such, we see here the necessity of soft constraints, even though for safety purposes, the penalty can be near infinite if violated, creating a virtual hard constraint.
The concept of the implementation is shown in the flowchart (Figure \ref{fig:flowchart}).

\noindent
This structure and the modus operandi used underline the need for a decision making system reducing the computational load, indeed, since we recompute the whole phase every fixed timestep, we need to make sure our system is efficient enough to reduce this time step as much as possible. 

\subsection{Results and discussion}

\begin{figure}
    \centering
    \begin{minipage}{.3\textwidth}
    \centering
    \includegraphics[width =1\textwidth]{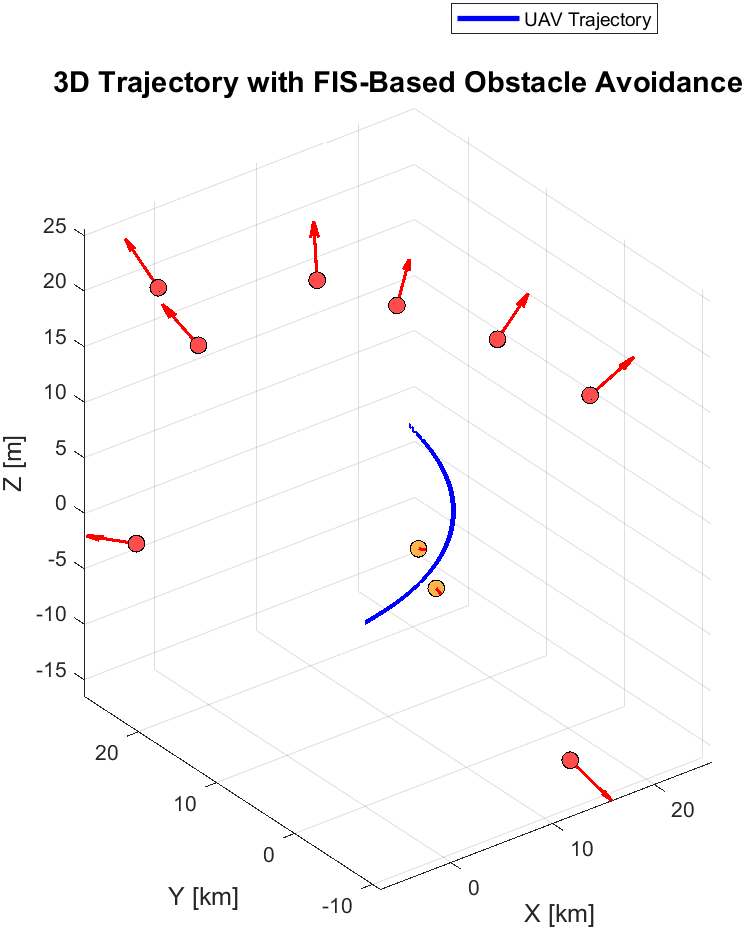}
    \caption{aircraft Trajectory and obstacles}
    \label{fig:traj}
    \end{minipage}
    \begin{minipage}{.3\textwidth}
    \centering
    \includegraphics[width = 1\textwidth]{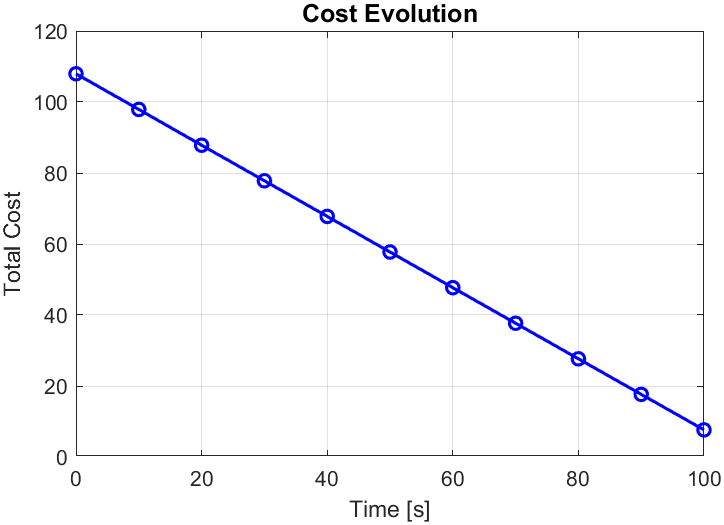}
    \caption{Cost of the optimization}
    \label{fig:cost}
    \end{minipage}
    \begin{minipage}{.3\textwidth}
    \centering
    \includegraphics[width = 1\textwidth]{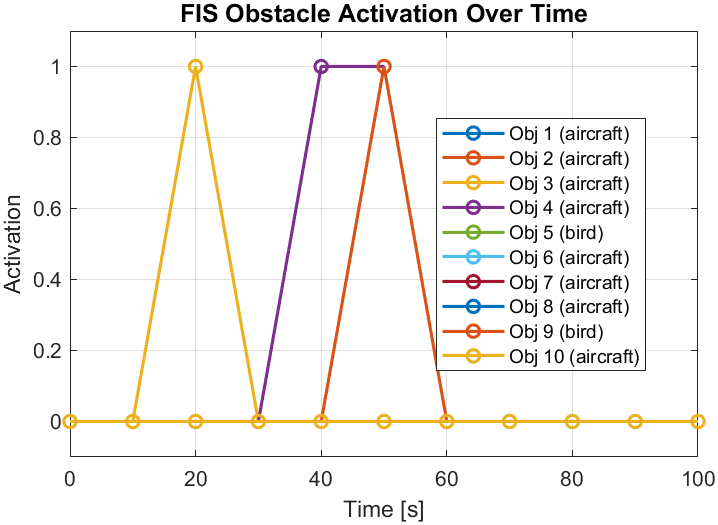}
    \caption{Obstacle activation}
    \label{fig:activation}
    \end{minipage}
    
\end{figure}

We first implemented the concept using a highly simplified aircraft model provided as an example within the FALCON toolbox. This initial abstraction, although not representative of realistic aircraft dynamics, served as a demonstrator to validate the methodology before transitioning to a more accurate state-space model of a different aircraft. Notably, even at this early stage, each optimization run required only 2–3 seconds in a single-threaded MATLAB environment, indicating promising potential for real-time or near real-time implementation once the framework is fully optimized.
After establishing this baseline, we integrated the fuzzy reasoning layer, resulting in the constraint-activation behavior illustrated in Fig. \ref{fig:activation}. Under these conditions, the solver produced a feasible optimal trajectory, however, further inspection revealed an unexpected issue. Specifically, the Lagrangian contribution to the cost function was identically zero throughout all simulations. As a result, obstacle-related constraints were effectively ignored by the optimizer, independent of the fuzzy layer’s decisions. This can be observed in Fig. \ref{fig:cost}, where the cost decreases linearly with the estimated time-to-arrival but shows no nonlinear variations or increases associated with Lagrange multiplier activity. Consequently, the resulting trajectories remained unchanged across runs, even when obstacle motions were varied (see Fig. \ref{fig:traj}).
To diagnose the problem, we examined the underlying optimization tools. The system relied on the FALCON toolbox (TUM) combined with IPOPT for nonlinear optimal control. When using the most recent versions of these components (IPOPT and FALCON v1.32), we consistently observed the degenerate zero-Lagrangian behavior described above. Such behavior is incompatible with the structure of the formulated Optimal Control Problem and therefore suggests a software level incompatibility or regression rather than an error in the modeling framework.

\section{Conclusion and Future Work}
Throughout this paper, we have presented a concept that is deeply rooted in real-world practice through its adherence to aviation regulations and air traffic management guidelines. The proposed optimal rerouting system is expected to operate with an update interval on the order of 2–3 seconds, based on the results obtained with the fully integrated FRBS and optimal control system. 
Under the assumption of the perfect radar, the FRBS is explicitly constructed in accordance with official separation standards and performs as intended, ensuring that its behavior is transparent, explainable, and directly traceable to regulatory requirements.
During the development of the system, we relied on the FALCON toolbox (TUM) together with IPOPT for nonlinear optimization. During integration of the latest releases of these software components, we encountered an unexpected incompatibility: the Lagrangian cost reported by the solver remained identically zero across all iterations, regardless of the problem configuration. This behavior is inconsistent with the structure of our optimal control formulation and therefore indicates a disruption in the solver–toolbox interaction rather than a modeling error.
To verify that the anomaly was attributable to software rather than to our methodology, we will revert to earlier versions of both FALCON and IPOPT to confirm that the issue is introduced by recent software changes rather than by the proposed FRBS decision making architecture in future work.
It is also expected to extend the study by adopting a higher fidelity model and integrating the FRBS with a Genetic Algorithm to optimize the membership functions, thereby enhancing overall system performance, guaranteeing output monotonicity, and improving resilience to noise.
Lastly, the robustness of the system to changing constraints will be assessed through Monte Carlo simulations, and the performance will be benchmarked against other anti-collision systems used at lower distances, such as Convolutional Neural Networks, full fuzzy controllers or reinforcement learning models.

\section*{Acknowledgments}

The authors extend their sincere gratitude to the members of the AI Bio Lab at the University of Cincinnati for their invaluable discussions and collaborative efforts that facilitated the realization of this work.

\newpage

%
%
%



\end{document}